\documentclass{article}

     \usepackage[nonatbib,final]{neurips_2020}

\usepackage[utf8]{inputenc} % allow utf-8 input
\usepackage[T1]{fontenc}    % use 8-bit T1 fonts
\usepackage{hyperref}       % hyperlinks
\usepackage{url}            % simple URL typesetting
\usepackage{booktabs}       % professional-quality tables
\usepackage{amsfonts}       % blackboard math symbols
\usepackage{nicefrac}       % compact symbols for 1/2, etc.
\usepackage{microtype}      % microtypography

\usepackage{graphicx}  % DO NOT CHANGE THIS
\usepackage{comment}
\usepackage{amsmath,amssymb} % define this before the line numbering.
\usepackage{color}

\DeclareRobustCommand{\eg}{e.g. }
\DeclareRobustCommand{\ie}{i.e. }
\DeclareRobustCommand{\etc}{etc. }

\def\dataset{`ApartmenTour' dataset }
\def\datasetEnd{`ApartmenTour' dataset}
\def\model{\emph{WAL }}
\def\modelatt{\emph{WAL-att }}
\def\modeladv{\emph{WAL-att-adv }}

\usepackage[symbol]{footmisc}

\title{\vspace{-0.1cm}Watch and Learn: Mapping Language and \\ Noisy Real-world Videos  with Self-supervision\vspace{-0.2cm}}

\author{
  Yujie Zhong \footnotemark \\
  Malong LLC \\
  \texttt{jaszhong@malongtech.com}
  \And
  Linhai Xie \\
   University of Oxford \\
   \texttt{linhai.xie@cs.ox.ac.uk} \\
  \And
  Sen Wang \\
  Heriot-Watt University \\
   \texttt{s.wang@hw.ac.uk} \\
  \And
  Lucia Specia \\
  Imperial College London \\
  \texttt{l.specia@ic.ac.uk} \\  
  \And
  Yishu Miao \\
  Imperial College London \\
  \texttt{ym713@ic.ac.uk} \\
}

\begin{document}

\maketitle

\addtocounter{footnote}{1}
\footnotetext[1]{Work done during PhD at Oxford}

\vspace{-8mm}
\begin{abstract}
\vspace{-4mm}

In this paper, we teach machines to understand visuals and natural language by learning the mapping between sentences and noisy video snippets without explicit annotations.
Firstly, we define a self-supervised learning framework that captures the cross-modal information.
A novel adversarial learning module is then introduced to explicitly handle the noises in the natural videos, where the subtitle sentences are not guaranteed to be strongly corresponded to the video snippets.
For training and evaluation, we contribute a new dataset `ApartmenTour' that contains a large number of online videos and subtitles. 
We carry out experiments on the bidirectional retrieval tasks between sentences and videos, and the results demonstrate that our proposed model achieves the state-of-the-art performance on both retrieval tasks and exceeds several strong baselines.

\vspace{-4mm}
\end{abstract}

\vspace{-2mm}
\section{Introduction}

\vspace{-4mm}

\begin{figure}[t]
\centering
\includegraphics[width=0.9\linewidth]{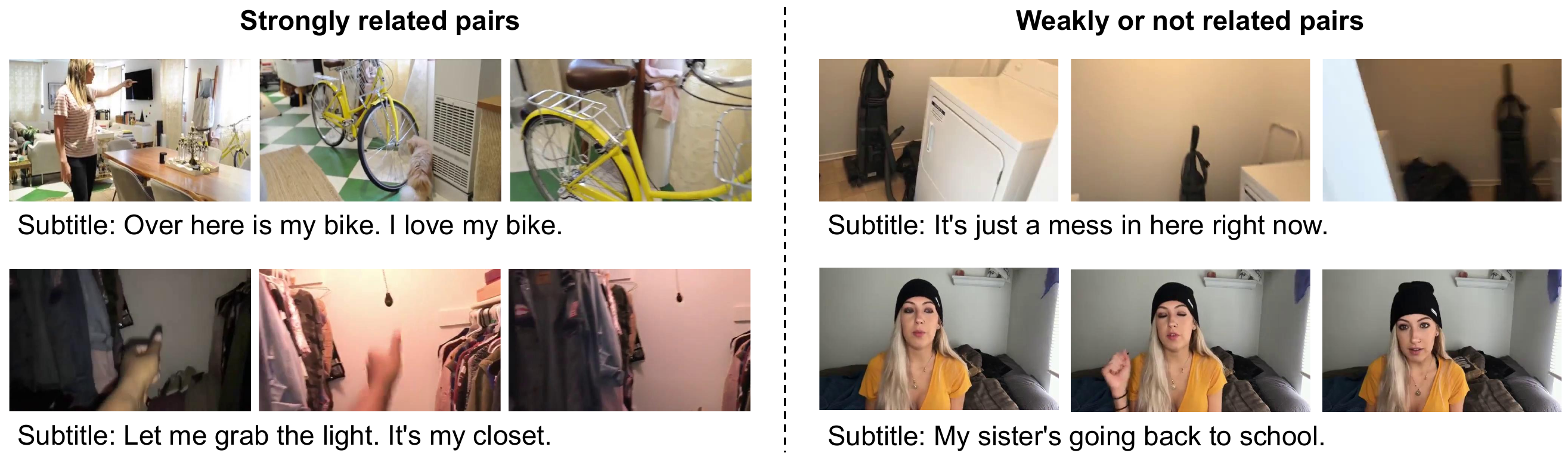}
\vspace{-2mm}
\caption{ {Examples of the \datasetEnd.} 
Three selected  frames  and the subtitle are shown for each video clip.
The top two rows are the strongly-related pairs; whereas the bottom ones can be seen as noises since they are weekly or not related pairs, which makes the learning very challenging.}
\vspace{-5mm}
\label{fig:teaser}
\end{figure}

Learning the mapping between vision and language in a supervised manner has been actively studied for many years~\cite{karpathy2014deep,kiros2014unifying,wang2016learning,lee2018stacked,wang2019camp,mahajan2019joint,burns2019language, sarafianos2019adversarial,ji2019saliency,wei2020multi,zhang2020context,wei2020universal,shao2018find,xu2018text,yu2018joint,chen2020fine}.
However, annotating parallel data costs a large amount of human labor (\eg ~\cite{lin2014microsoft,johnson2016densecap}).
Therefore, we propose to straightforwardly teach machines to understand online videos with transcribed subtitles from speech.
Certainly, the subtitles are not the perfect description, but they are naturally aligned with the videos, and a lot of them are explicitly or implicitly grounded by the objects and scenes from the videos.
Following the intuition of how humans learn to pick up new knowledge from the real-world, the machine learns to selectively choose the pairs and learn the mapping of the visual and language.
In this case, we have access to massive unlabelled video datasets online for visual and language grounding. % by a simple watch-and-learn process.
We collect a large video dataset without annotations, termed the \datasetEnd.
These videos are decent source materials for capturing the language-visual correspondence (LVC), since most of them are describing the indoor scenes, objects or surroundings (see Figure~\ref{fig:teaser}).

However, learning the LVC on these noisy videos is very challenging, because there is still a lot of subtitles and video frames that are loosely correlated, unlike the parallel dataset of image captioning (\eg MS-COCO) where the captions are produced by humans to describe the image on purpose. 
For instance, the author may talk about things irrelevant to the scenes (\eg the bottom-right example in Figure~\ref{fig:teaser}). 
In addition, the videos provide sequential visual signals instead of a single image with caption, and the sentences have to be mapped over consecutive frames to locate the corresponding ones in a certain video clip, which further complicates the cross-modal learning in this case.
Therefore, simply applying encoder-decoder models is not good enough to tackle this problem.

In this work, we propose a framework for learning the LVC by minimising a cross-entropy loss that brings together the subtitle and video pairs.
An attention mechanism is employed to dynamically focus on the frames of a video clip that correspond to the subtitle sentence and disregard the irrelevant ones. 
In addition, we apply an adversarial loss during training such that the network is able to selectively start with learning from simple examples (\ie video-sentence pairs with obvious correspondence), and then gradually move on to the difficult ones.
The experiments quantitatively and qualitatively demonstrate the superiority of the proposed model compared to the strong baselines, and show the great potential of watch-and-learn strategy.

\vspace{-3mm}
\paragraph{Related work.} 
A related work to this paper is~\cite{miech2020end} which aims at learning the LVC from a large corpus of narrative videos~\cite{miech2019howto100m} with self-supervision. While~\cite{miech2020end} focuses on designing a novel loss function, we propose to dynamically choose training samples based on the learning difficulty.

\vspace{-3mm}
\section{Methodology}
\label{sec:lvc}

\vspace{-2mm}
\subsection{Problem Definition}
\vspace{-3mm}

\paragraph{Language-visual correspondence.} %\label{sec:LVC}

The essence is to leverage the fact that a lot of visual scenes, objects and even actions are naturally aligned with the speech in the online videos.  
The LVC task on the collected \dataset (more details of the dataset is described in Section~\ref{exp}) can be very challenging due to several reasons. 
First, there is a significant amount of time in each video when the author is not describing the apartment. 
Second, there also exists a lot of cases where the visuals only contain a small subset of the things described in the speech
or only a small portion of the frames corresponds to what the subtitle refers to.
The existence of loose-correspondence can severely confuse the system in the training process as they introduce some noises.

\vspace{-3mm}
\paragraph{Loose-correspondence alleviation.} %\label{sec:adv}

Intuitively, it is easier for the network to learn the language-visual correspondence from the training pairs with relatively obvious correspondence at the beginning of the training, and leave out the less obvious ones which can be involved to the training at later stages.
This training scheme shares a similar spirit with the curriculum learning~\cite{bengio2009curriculum} which presents training examples in a meaningful order, \ie gradually illustrating more complex ones. 
It is shown to improve the performance of networks in many machine learning tasks, especially in weakly supervised learning~\cite{gong2016multi,guo2018curriculumnet}. 
Following this intuition, we propose a network named Watch-And-Learn (\model) to alleviate this loose-correspondence.

\vspace{-3mm}

\subsection{Network Architecture}

\vspace{-2mm}
\paragraph{Vision channel and language channel.}
The vision channel in the \model network extracts and embeds the visual representation of videos with the frame-level attention. 
\begin{figure*}[t]
\centering
\includegraphics[width=0.85\linewidth]{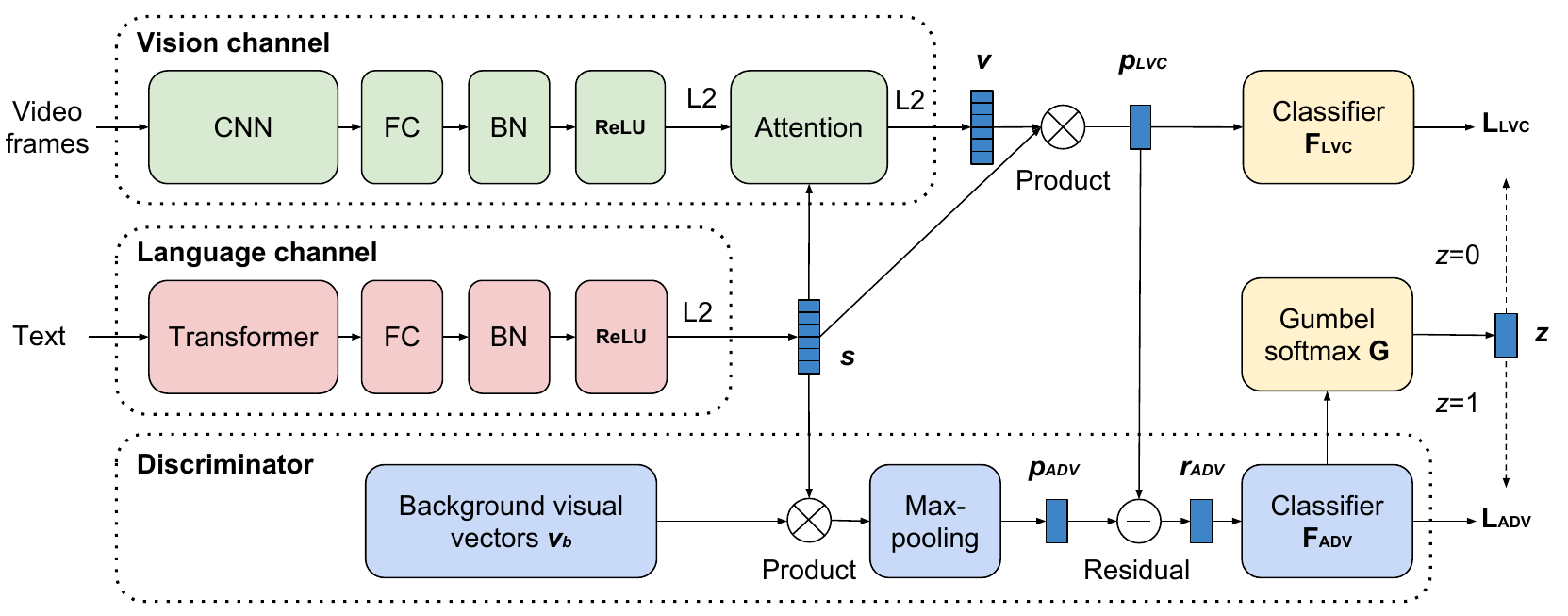}
\vspace{-3mm}
\caption{ {Network architecture of \model.} 
   The proposed network architecture consists of three channels.
   The {vision channel} and the {language channel} has an encoder network (ResNet-34 and BERT) and an embedding module that consists of a FC layer, a batch-norm layer and a ReLU layer. The vision channel has an additional attention module which enables the sentence feature $s$ to attend to the related frames in the video.
   The {discriminator}  consists of a stack of background visual vectors, a max-pooling layer and a classifier. The aim of the discriminator is to dynamically determine whether an input sentence-video pair is suitable for the LVC training at the current state. The binary decision is obtained by sampling a discrete latent variable $z$ using a gumbel-softmax layer.
   The input pair contributes to the LVC loss if $z$ is 0, otherwise it contributes to an adversarial loss.}
\label{fig:network}
\vspace{-5mm}
\end{figure*}
We experiment with three different attention mechanisms in the attention module, namely dot-product~\cite{luong2015effective}, multiplicative~\cite{luong2015effective} and additive~\cite{bahdanau2014neural}. %They can be expressed as equation~\ref{eq:att-dot}, \ref{eq:att-mul} and~\ref{eq:att-add}, respectively.
The language channel consists of a 12-layer bidirectional transformer~\cite{devlin2018bert} and an embedding module. 
The embedding module embeds the BERT representation to a common space with the embedded visual features, giving $s$ in Figure~\ref{fig:network}.

\vspace{-3mm}
\paragraph{Discriminator.}
The aim of this discriminator (bottom row in Figure~\ref{fig:network}) is to dynamically determine whether a sentence-video pair is suitable for training or not at the current state.
If a pair is considered not suitable for the training at the current state, then this pair does not contribute to the LVC objective. Instead, it contributes to an adversarial loss function $L_{\text{ADV}}$.
We interpret how corresponded or informative a sentence-video pair is by the use of some background visual vectors. Acting as inductive biases, the introduced background visual vectors can be seen as some abstracted features that directly capture some background semantics or information which are difficult to interpret visually. The background visual features are randomly initialized and jointly learned with \model. More details refer to the supplementary material.
The intuition of the max-pooling layer is to find the largest similarity between the input sentence feature $s$ and all of the background visual features $p_{\text{ADV}}$. 
To determine whether a sentence-video pair is beneficial for the training, we compare the similarity between the input sentence-video pair $p_{\text{LVC}}$ with $p_{\text{ADV}}$. If $p_{\text{ADV}}$ is larger than $p_{\text{LVC}}$ by a threshold (which could be either negative or positive), we then tend to consider the input pair to be not suitable for the training at the current state. However, this input pair might come into play at later training period when the network becomes more capable.
The discriminator channel adopts an adversarial training loss for learning, which is explained in Section~\ref{sec:loss}.

\vspace{-3mm}
\paragraph{Relation to existing methods.}
Common cross-modal retrieval models require human annotations as explicit learning signals. In contrast, the proposed WAL network is designed to learn from noisy data in real-world scenarios. As one of the key contributions in this work, the discriminator enables the network to dynamically discard the noise and learn from the meaningful sentence-video pairs.

\vspace{-4mm}
\subsection{Training Losses}\label{sec:loss}
\vspace{-2mm}
The training loss $L_{overall}$ is the sum of two binary cross-entropy losses, $L_{\text{LVC}}$ and $L_{\text{ADV}}$.
$L_{\text{LVC}}$ corresponds to the LVC task, \ie determining whether a sentence-video pair is matched. The positive pairs are naturally obtained by time alignment. We then utilize the negative sampling method to sample the negative pairs. 
The general loss function for the $j$th pair can be expressed as Equation~\ref{eq:loss1}, where $y_j$ denotes the label for the $j$th training pair, $\sigma$ is a sigmoid function. $y_j$ can be either 0 (unmatched) or 1 (matched).
\vspace{-1mm}
\begin{flalign}
\begin{split}
    L(j, f_j) = y_j \log(\sigma(f_j)) + (1-y_j) \log(1-\sigma(f_j)).
    \label{eq:loss1}
\end{split}
\vspace{-2mm}
\end{flalign}
When $z=0$, we have $L_{\text{LVC}} = L(j, F_{\text{LVC}}(s_j^\top  v_j))$, where $F_{\text{LVC}}$ is the classifier applied on top of the dot product between the sentence feature $s_j$ and the video feature $v_j$. 
When $z=1$, the loss $L_{\text{ADV}} = L(j, F_{\text{ADV}}(\psi (s_j^\top v_b) - s_j^\top v_j))$ applies instead, where $F_{\text{ADV}}$ is the classifier of the discriminator, $\psi$ is max-pooling operation. 
In particular, we fix $y_j$ for any training pair to be 0 for $L_{\text{ADV}}$ to encourage the network to gradually involve more training pairs for $L_{\text{LVC}}$. Hence, $L_{\text{ADV}}$ acts as an adversarial loss.
The overall loss is therefore the sum of the individual loss over all the pairs.

\vspace{-3mm}
\section{Experiments}\label{exp}

\vspace{-1mm}

\vspace{-1mm}
\paragraph{Dataset.} The collected \dataset contains 2906 YouTube videos in a resolution of 480P. The average length of the videos is about 12 minutes. Each video comes with subtitles provided by either the author or auto-generated by YouTube. In total, there are around 500k sentence-video pairs extracted from this dataset.

\vspace{-3mm}
\paragraph{Evaluation protocol.}
We use two standard evaluation metrics for retrieval -- mean average precision (mAP) and recall at position $k$ (Rec@k). mAP measures the precision of the predicted ranking averaged over all positions, while rec@k measures how well the positives are retrieved at position $k$. In this following experiments, we set $k$ to be 5 (and 10 for the second task). 
Both metrics are multiplied by 100 such that the reported values range from 0 to 100.

\vspace{-3mm}
\paragraph{Baselines.}
We compare our networks to 7 baseline methods. 
Firstly, we report the performance of random guesses as a reference, indicating how much the networks learn from the proposed self-supervision task.
The next three baselines formulate the retrieval task as a generative modelling problem, and adopt the Neural Image Caption (NIC) architecture introduced in~\cite{vinyals2015show}. 
The three baselines differ in the training procedure.
We further propose \emph{NIC-BERT} baseline that adapts the current NIC network for the retrieval task at hand, by mapping both videos and sentences to language features using BERT.
Lastly, some strong baselines are implemented. SCAN~\cite{lee2018stacked}, PVSE~\cite{song2019polysemous} are typical supervised learning methods and achieved state-of-the-art performance in the text-image mapping task on MS-COCO.

\vspace{-3mm}
\subsection{Sentence-video bidirectional retrieval}
\label{sec:test1}
\vspace{-2mm}

\begin{table}[t]
\centering
\fontsize{8}{10}\selectfont
\caption{{Performance on the \datasetEnd.} 
Video search denotes querying videos using sentences, and vice versa for sentence search.
Rec@5 is the recall at position 5. Higher is better for both mAP and Rec@5. 
`AT' and `MS' denote the \dataset and MS-COCO~\cite{lin2014microsoft}.}
\vspace{-2mm}
\setlength{\tabcolsep}{1mm}
\begin{tabular}{cc|cccc}
\hline
Network  & Training            & \multicolumn{2}{c}{Video Search} & \multicolumn{2}{c}{Sentence Search}  \\
         &        Data                   & mAP & Rec@5           & mAP              & Rec@5          \\ \hline
Random   & -                        & 5.2               &  5.0               &  5.2                &  5.0             \\
NIC-BERT & MS                  & 7.8            &   12.0              & 6.1              &        9.0       \\ 
NIC      & MS                  & 19.7           &  30.0               & 22.6             &       37.0        \\
NIC      & AT           & 16.9           &  25.0               & 18.8             &        26.0     \\
NIC      & MS+AT & 25.7           &   36.0              & 25.7             &       41.0       \\ 
SCAN LSE+AVG    & AT        & 28.1               &  41.3               & 28.2          & 42.6       \\ 
PVSE & AT & $27.5$ & $40.3$ & 27.1 & 42.4        \\

\hline
WAL      & AT           & 25.0           &  39.0               & 26.3                 &      39.8        \\
WAL-att  & AT           & 27.3           &  41.6               &  27.9        & 42.0 \\ 
WAL-att-adv  & AT     & \textbf{30.1}           & \textbf{44.2}           &  \textbf{30.0}          & \textbf{45.4}   \\

\hline
\end{tabular}

\label{tab:test1}
\vspace{-5mm}
\end{table}

\paragraph{Test set.}
The dataset for testing consists of 100 positive sentence-video pairs. Each video is of length 2 to 5 seconds. 
Unlike the training set, the positive pairs in the test and validation set are manually annotated and guaranteed to be \emph{strongly} related.
We query the 100 videos using each sentence and average the results for text-video retrieval, and vice versa for the video-sentence search.

\vspace{-3mm}
\paragraph{Comparison with existing methods.}
As Table~\ref{tab:test1} shows, \modeladv outperforms all the baselines by a large margin.
A surprising observation is that even \model performs on par with the strongest baseline \emph{NIC} which is trained on MS-COCO and fine-tuned on the \datasetEnd. Note that training on MS-COCO is fully supervised and thus provides a very good pre-training.
By comparing the three \emph{NIC} baselines, we see that the model trained on MS-COCO can be directly applied to the retrieval task off-the-shelf and achieve competitive results. The performance can be further boosted by fine-tuning on the \dataset as a way to adapt the model to the test set. However, simply training on the \dataset in a generative manner does not provide strong results. This could be due to the noises in the dataset (\eg loose-correspondence) and incomplete sentences \etc
The performance \emph{NIC-BERT} is far from other BERT methods. This is because the generated captions by NIC is quite different from the videos subtitles, \eg the NIC trained on MS-COCO tends to focus on the person in videos while it is not the case for the subtitles in the \datasetEnd. 

SCAN and PVSE are the strongest baselines, as we expected. By combining two complimentary formulations of attention, SCAN performs similar to \emph{WAL-att}, but worse than \emph{WAL-att-adv}. 
Similarly, PVSE performs on par with SCAN, but is outperformed by \emph{WAL-att-adv}.
This further proves the significance of using the discriminator when training networks on a noisy dataset.
Moreover, our method is much more memory-efficient (\ie no need to keep detection features).
In general, all the networks trained on the \dataset work significantly better than random guesses. This demonstrates the value of the dataset and the effectiveness of the proposed LVC learning task.

\vspace{-3mm}
\paragraph{Discriminator.}
It is impressive to see that the discriminator channel enhances the performance of \modelatt significantly, \eg $27.3$ to $30.1$ in mAP
for image search.
This result demonstrates the benefit of the proposed discriminator channel during training. 
At the beginning of the training, about 60\% training pairs contribute to the LVC learning; this number gradually increases and then stabilizes at 95\% at the late training epochs, meaning that most of the pairs take part in the LVC training. More details on the behaviour of the discriminator during training refer to the supplementary materials.

\vspace{-4mm}
\section{Conclusion}\label{conclusion}
\vspace{-3mm}
We define a novel learning task based on the intuition of the language-visual correspondence,
and propose a network architecture that can be trained effectively for cross-modal retrieval between sentences and videos,  
which is shown by both quantitative and qualitative experiments.

%\bibliographystyle{plain}
%\bibliography{mybib}

%\clearpage

\section{Appendix}

\subsection{Additional Details for the Proposed Method}

\paragraph{Background visual features.}
We interpret how corresponded or informative a sentence-video pair is by the use of some background visual features. 
Acting as inductive biases, the introduced background visual features can be seen as some abstracted features that directly capture some background semantics or information which are difficult to interpret visually.
The input sentences without much visual semantics are expected to have high similarities with them, which contribute little to the LVC task.
Therefore, we compare the similarity between the input sentence and the background visual features to that between the input sentence and the input video clip, as a way to determine how corresponded or visually informative an input sentence-video pair is. 
The background visual features are randomly initialized and jointly learned with the WAL network.
We use 4 background image features in the discriminator channel, which are initialized by random grouping of the L2-normalized video features $v$ in the training set extracted using the pre-trained ResNet-34 and the subsequent randomly initialized FC layer in the vision channel.

\paragraph{Adversarial training.}
To enable the network to gradually learn more complex concepts, the LVC objective can be learned in an adversarial manner. To achieve this, we fix the label $y_j$ for any training pair to be 0 in the discriminator for the $L_{\text{ADV}}$. This is because we expect to train the network with more pairs (especially the difficult ones) gradually, and fixing $y_j$ to be 0 reduces the probability of sampling the $z$ to be 1 (\ie not suitable for training) along the training.
This, therefore, can be considered as an adversarial training. Namely, during training, the network tends to sample all the pairs to contribute to the $L_{\text{ADV}}$ (\ie $z=1$), since in the discriminator the network only needs to always predict 0 (recall that all the labels $y$ for $L_{\text{ADV}}$ is fixed to 0).

\vspace{-3mm}
\subsection{Dataset Collection}
\label{supp:data}

To obtain the urls of the videos, we use `apartment tour' and its two extensions (by adding a year or a city name) as the query string query the YouTube search engine.
We range the year from 2008 to 2018, and use 28 cities in the world.
We manage to download 2906 videos with subtitles.
we cut the videos into clips based on the subtitles and corresponding timing. Only the clips with human speech are preserved, while those with no speech or not useful subtitles such as sound indicators are removed. Each clip lasts about 2 to 5 seconds, and 2 frames per second are extracted from each clip. In total, we obtain over 500k pairs.

\subsection{Implementation Details}
\label{supp:train}

\paragraph{Training details for WAL networks.}
For each mini-batch during training, we randomly sample the same number of positive and negative sentence-video pairs to avoid the class-imbalance problem. The negative pairs are obtained by first sampling a video, and then randomly sampling a sentence from the \dataset, excluding the original paired sentence.
For each epoch, we ensure that every video clip in the dataset is used. During training, we sample a fixed number $N_f$ of frames for each the video for efficiency and as a way of data jittering. If a video has less than $N_f$ frames, some frames are reused. We set $N_F=5$ in our experiments. 
Throughout the training, we keep the weights of the pre-trained BERT model and ResNet-34 fixed, as a way to prevent overfitting. 
Stochastic gradient descent is used to train the network, with weight decay 0.001, momentum 0.9, and an initial learning rate of 0.1; the learning rates are divided by 10 in later epochs.
In particular, for the first 50 epochs, the discriminator channel is fixed and its gradients are stopped. This helps to reduce the variance of the gradients at the early training epochs~\cite{wen2017latent}. The whole network is then trained for 20 more epochs.

\paragraph{Data augmentation.}
The training frames are obtained as follow: first, a crop of a random size (ranging from $0.8$ to $1$) and a random aspect ratio (between $3/4$ and $4/3$) of the original image is taken. The image crop is then resized to $224 \times 224$. A random horizontal flipping and mean subtraction is applied to the image crop. The resultant image is then used as the input to the visual channel of the network.
At test time, images are resized so that the smallest dimension is 256
and the central crop of $224 \times 224$ is taken.

\paragraph{Training details for NIC baselines.}
The architecture of NIC baselines consists of a ResNet-152~\cite{he2016deep} for image feature extraction and a LSTM for caption generation conditioned on the feature of the input image. 
The ResNet-152 is pre-trained on ImageNet~\cite{deng2009imagenet} and fixed for further caption generation training except for batch-norm layers. 
The training of the NIC baselines is mainly on the sentence decoding part.
In detail, the output of a ResNet-152~\cite{he2016deep} is first embedded into a feature of size 256 before feeding to the LSTM. The size of the hidden state of the one-layer LSTM is 512. During  training for image captioning, the weights of the LSTM, the embedding layer and all the batch-norm layers in the ResNet-152 are updated. During training on the \datasetEnd, for each sentence-video pair, a frame is randomly selected from the video as the image input.

\paragraph{Retrieval method for NIC baselines.}
At test time, we use the same ranking method as described in~\cite{mao2014explain}. For the sentence-video retrieval, we measure the perplexity of generating the query sentence given each dataset video, which essentially represents how likely each frame in each dataset video can generate the query sentence. 
The perplexity of a sentence
can be expressed as $ \log PPL(w_{1:L}|I) = -\frac{1}{L}\sum_{n=1}^{L}\log P(w_n|w_{1:n-1},I)$,
where $L$ is the length of the sentence, $P(w_n|w_{1:n-1},I)$ is the probability of generating the word $w_n$ given image $I$ and
previous words $w_{1:n-1}$.
The perplexity of each video is the average perplexity of its frames.
Whereas for the video-sentence retrieval, we measure the perplexity of generating each dataset sentence given the query video. Note that the perplexity of each sentence is normalized in the way described in~\cite{mao2014explain} to reduce the bias of the sentence frequency.
The training procedure of these three baselines is in the supplementary material.

\paragraph{Retrieval method for NIC-BERT baseline.}
For the videos, we first use NIC to generate a caption for each frame. 
In this way, each video is represented by a collection of sentences. 
We then use BERT to extract language features from the caption of each video frame to form the final video representation.
Similarly, the sentences are also represented by their BERT features. As a result, the retrieval can be performed by similarity computation between the BERT features of the videos and the sentences.

\paragraph{Training details for SCAN.}
To train SCAN~\cite{lee2018stacked}, we first use a Faster-RCNN~\cite{ren2015faster} (with ResNet-101) trained on Visual Genome~\cite{krishna2017visual} to extract the features of 36 ROIs with highest confidence scores in each frame in the dataset. SCAN is then trained on the pre-computed features for 30 epochs with the same $\lambda$ values as in the paper. For both training and testing SCAN, the similarity score between a sentence and a video is the maximum score between the sentence and all frames. 

\paragraph{Training details for PVSE.}
PVSE~\cite{song2019polysemous} is trained on the \dataset with the weights of the diversity loss and the domain discrepancy loss being 0.1 and 0.01. Pre-trained ResNet-152 is used for visual feature extraction and GloVe~\cite{pennington2014glove} is used for word embedding. The maximum number of video length of 5. The number of local embeddings used is 3. The network is trained on the \dataset for 70 epochs.

\iffalse
\subsection{Time interval localization Details}

\paragraph{Localization by Retrieval.}
The time interval localization task is formulated as a retrieval problem in our experiments.
Namely, each video is split into short clips of around 3 seconds, and the model needs to rank all the clips in order to localize the time interval corresponding to the text query. We therefore use the retrieval evaluation metrics to measure 
how well the target clips are retrieved by the model, %as a reflect of 
the time localization ability.
Notably, different from the bidirectional retrieval task, the query in this task may correspond to more than one consecutive clip in the video.

\paragraph{Test set.}
The test set consists of 20 full videos randomly sampled from the \dataset (exclusive to the training set).
 On average, each video has a length of 12 minutes and about 240 short clips. 
There are 63 query sentences in total (more than 3 queries per full video on average). 
The queries are selected by human after watching the full videos. They should satisfy two selection rules: the sentence and the video clip should match with each other, and the clip should be unique to the query sentence in the full video.
The results are averaged over the 63 queries.
\fi

\subsection{More Results and Visualization}

\subsubsection{Ablation Study}

\begin{table}[t!]
\centering
\fontsize{8}{10}\selectfont
\caption{{Ablation study on the sentence-video bidirectional retrieval.}
`BVF' denotes the number of background visual features. $G$ is the sampling layer of $z$ and `Soft.' denotes softmax. `In.' denotes the input to the classifier $F_\text{ADV}$ in the discriminator. `Con.' means the concatenation of $p_\text{LVC}$ and $p_\text{ADV}$.}
\vspace{-2mm}
\begin{tabular}{lcccc}
\hline
Network   & \multicolumn{2}{c}{Video search} & \multicolumn{2}{c}{Sentence search} \\
         &  mAP            & Rec@5           & mAP              & Rec@5            \\ \hline
         
WAL-att (BVF=$4$)   & 27.3           &  41.6               &  27.9        & 42.0 \\ 
WAL-att-adv (BVF=$4$)      & \textbf{30.1}           & 44.2           &  30.0          & 45.4   \\\hline
WAL-att-adv (BVF=$16$)       & 29.8       & 44.0            & \textbf{30.3} & \textbf{45.8} \\
WAL-att-adv (BVF=$64$)        & 30.0         & 44.1               & 30.2        & 45.5 \\  
WAL-att-adv ($\textbf{G}$=Soft.)    & 28.8          & 43.2              & 28.9       & 44.2  \\
WAL-att-adv (In.=Con.)       & 30.0          & \textbf{44.5}               & 30.0       & 45.2 \\
WAL-att-adv (In.=$p_\text{ADV}$)       & 28.7          & 43.0               & 29.0    & 44.5 \\ \hline

WAL-att (Triplet)          & 26.7               & 40.3                 & 26.9          & 41.2                 \\
WAL-att-adv (Triplet)     & 28.9          & 43.6         & 29.2       & 44.1              \\ 
\hline
\end{tabular}
\label{tab:ablation}
\vspace{-2mm}
\end{table}

We first conduct ablation studies on the discriminator, including varying the number of background visual features (BVF), the sampling layer and the input to the classifier $F_\text{ADV}$. We also compare two losses: the cross-entropy loss (introduced in Section~\ref{sec:loss}) and the commonly-used triplet loss.
The results are shown in Table~\ref{tab:ablation}. 
First, 
the performance is not very sensitive to the number of BVF in the discriminator,
\ie both models (\ie BVF=16 and 64) achieve similar performance as the model with BVF=4.
In particular, when BVF applies a large number (\eg 64), we observe that fact that many BVF never fire during training.
This can explain the insensitivity of the model to high values of BVF.
Second, replacing the gumbel-softmax with the standard softmax also enhances the performance, by comparing \emph{WAL-att-adv} ($G$=soft.) with \emph{WAL-att}.
But the improvement is far smaller than that using the variational inference with a discrete latent variable. This is probably because the training with variational inference and the discrete latent variable is closer to the binary decision that human would make, while the softmax operation is not as intuitive.
Third, we can conclude that feeding both $p_\text{ADV}$ and $p_\text{LVC}$ (in way of either residual or concatenation) to the classifier $F_\text{ADV}$ in the discriminator is better than feeding $p_\text{ADV}$ only. Intuitively, by comparing the two similarity scores ($p_\text{ADV}$ and $p_\text{LVC}$), the discriminator can dynamically make the decisions on whether a sentence-video pair is beneficial to the current training.
Lastly, we adapt the proposed architecture to train with the triplet loss.
Namely, the similarity of each positive sentence-video pair should be larger than those of two negative pairs (\ie one is the positive clip and the hardest negative sentence in a mini-batch, and vice versa for the other) by at least a margin (0.2).  
As Table~\ref{tab:ablation} (bottom part) shows, the triplet loss performs on a par with (marginally worse than) the LVC loss. We think that the hardest negative sampling might be too aggressive for such a noisy dataset with many false positive alignments.

\subsubsection{Discriminator Behaviour}
\label{supp:discriminator}

\begin{figure}[t]
\centering
\includegraphics[width=0.69\linewidth]{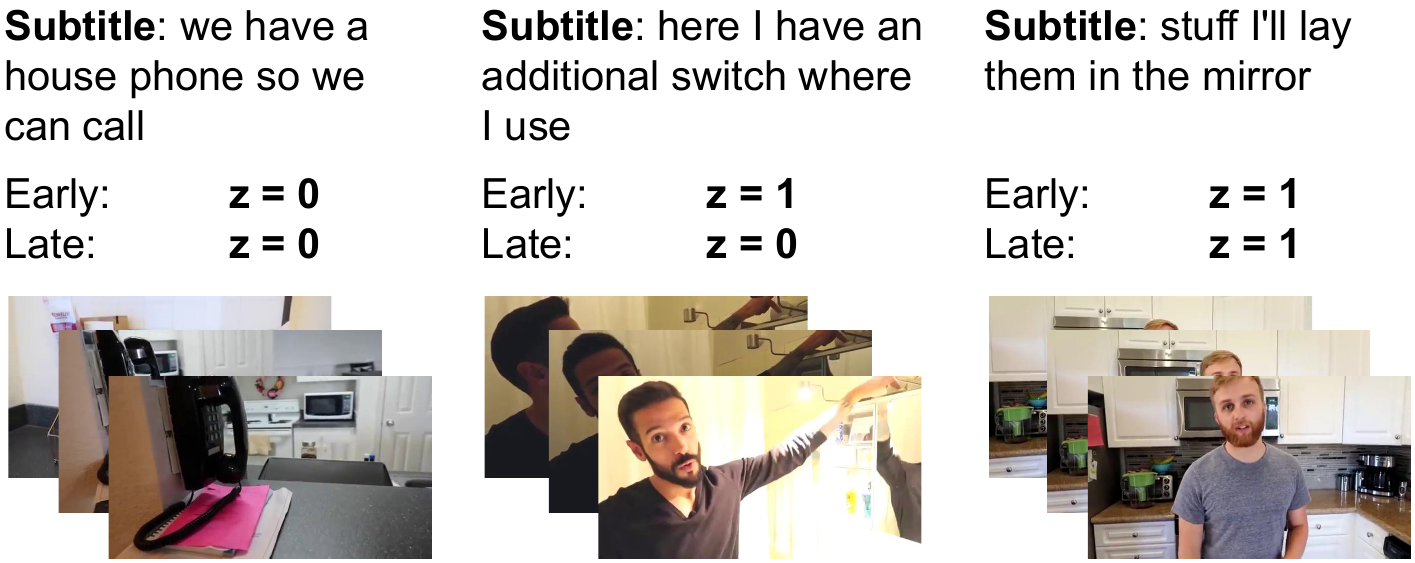}
\caption{ {Examples of the discriminator sampling during training.}
$z$ is shown above each video for the early and late training stages.
The first pair is always used for training as the phone is obvious.
The second example is not so obvious to determine if they are matched, as it requires the network to infer the relation between the switch and the change of the lighting condition in the video. Hence it is used for the LVC only in the late stage.
The last pair is not used for the LVC objective throughout the training, as it is a false positive.
}
\label{fig:gumbel}
\vspace{-4mm}
\end{figure}

Figure~\ref{fig:gumbel} shows some examples of the sentence-video pairs that are sampled to the adversarial loss (\ie $z=1$) during training, meaning that they are considered not suitable for the LVC training. The first two examples are sampled at the early training stages while the other two are sampled at the later training stages. We can see that the network gradually learns from more difficult pairs along the training. However, some pairs are not informative for the learning even in the late training stages.

\begin{figure}[t]
\centering
\includegraphics[width=0.6\linewidth]{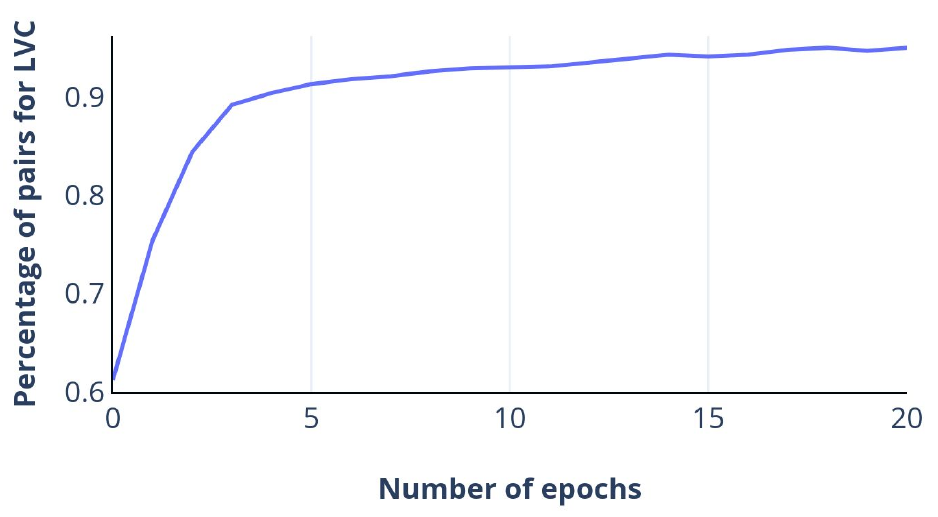}
\caption{ { The behaviour of the discriminator during training.} 
The horizontal axis is the number of epochs that the training takes,
and the vertical axis denotes the percentage of sentence-video pairs that correspond to $z=0$.
Each epoch covers all the video-sentence pairs in the training set.}
\label{fig:gumbel_plot}
\vspace{-2mm}
\end{figure}

During the training of the network, we record the number of sentence-video pairs that are sampled to contribute to the language-visual correspondence (LVC) learning task (\ie the discrete latent variable $z$ is sampled to be 0) in each batch. Furthermore, to reduce the variance, we take the average over all the numbers in each epoch. The resultant numbers are then divided by the batch size, which is 60 in our experiments, to give the percentage of the training pairs that correspond to $z=0$. 
Note that the rest of the pairs in each batch contribute to the adversarial loss.

Figure~\ref{fig:gumbel_plot} shows how the percentage changes along the training. Note that  we only show the percentage of the last 20 epochs of the training, in which the discriminator is trained together with the rest of the network. 
As we can see in Figure~\ref{fig:gumbel_plot}, the percentage is around 60\% before the training starts (\ie the 0th epoch). It then reaches 75\% after one epoch, meaning that $1/4$ of the sentence-video pairs do not take part in the LVC learning at this point. The percentage increases quickly in the first few epochs to about 90\% and then grows slowly to 95\% as the training proceeds.
This trend matches our expectation that the discriminator gradually allows more and more sentence-video pairs to contribute to the LVC loss, by the novel design of an adversarial training setting. Moreover, some pairs are still not used for the LVC learning task at the very end of the training, \eg the false positive pairs.

\subsubsection{Visual Attention Examples}
\label{supp:example}

\begin{figure*}[!t]
\centering
\includegraphics[width=0.7\textwidth]{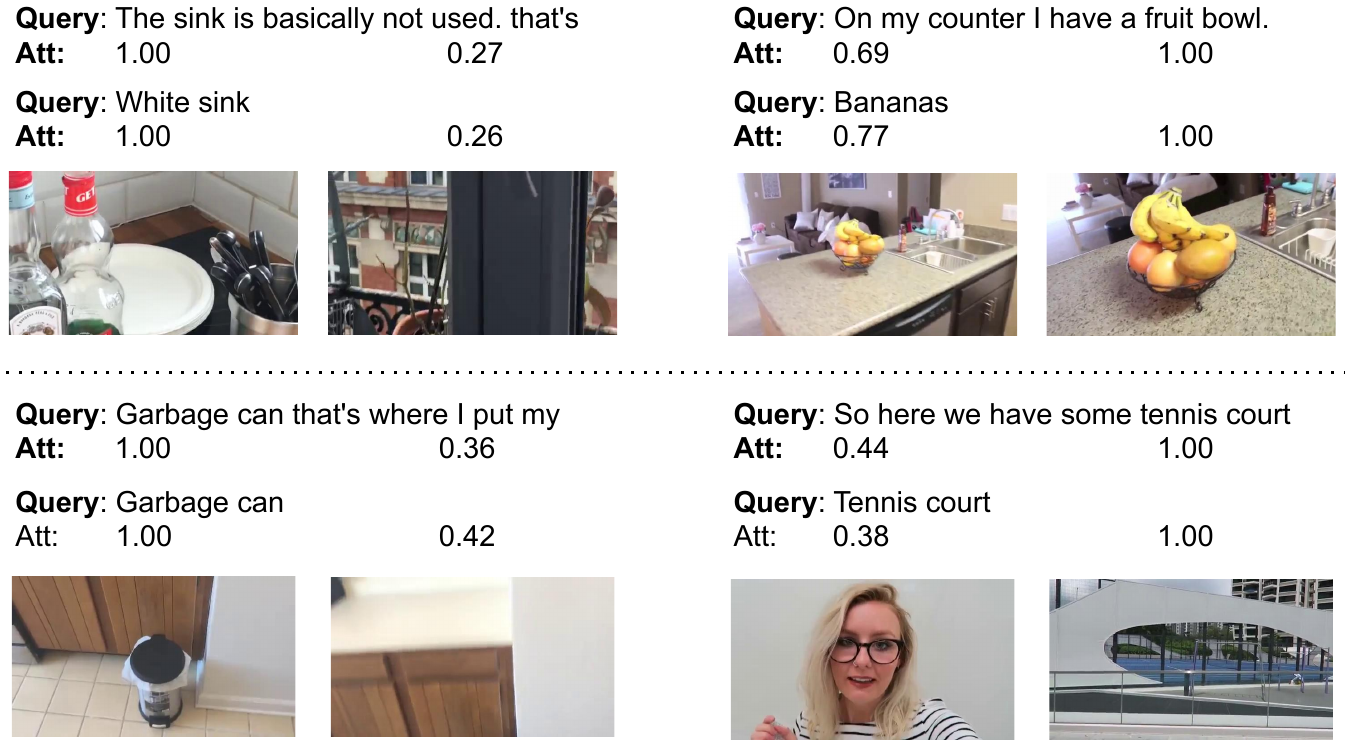}
\caption{ { Examples of frame-level attention} by query sentences. 
Each row contains two sentence-video pairs. In each example, the first query is the subtitle corresponding to the video clip (of which only two representative frames are displayed for better viewing), and the second query is simply a phrase or word.
The attention score, denoting how much the frame contributes to the final video feature, is shown above each frame.
Note that the absolute value is related to the number of frames in the video.
For better comparison, the attention scores are normalized such that the larger score is 1.
}
\label{fig:attention}
\end{figure*}

Some examples of the visual attention are shown in Figure~\ref{fig:attention}. It is  interesting to find that the attention mechanism also applies to simple query phrases or words (even that do not appear in the original subtitle as in the first row). This proves that the learnt representations do not overfit to the original subtitles.

\begin{figure*}[t!]
\centering
\includegraphics[width=0.65\linewidth]{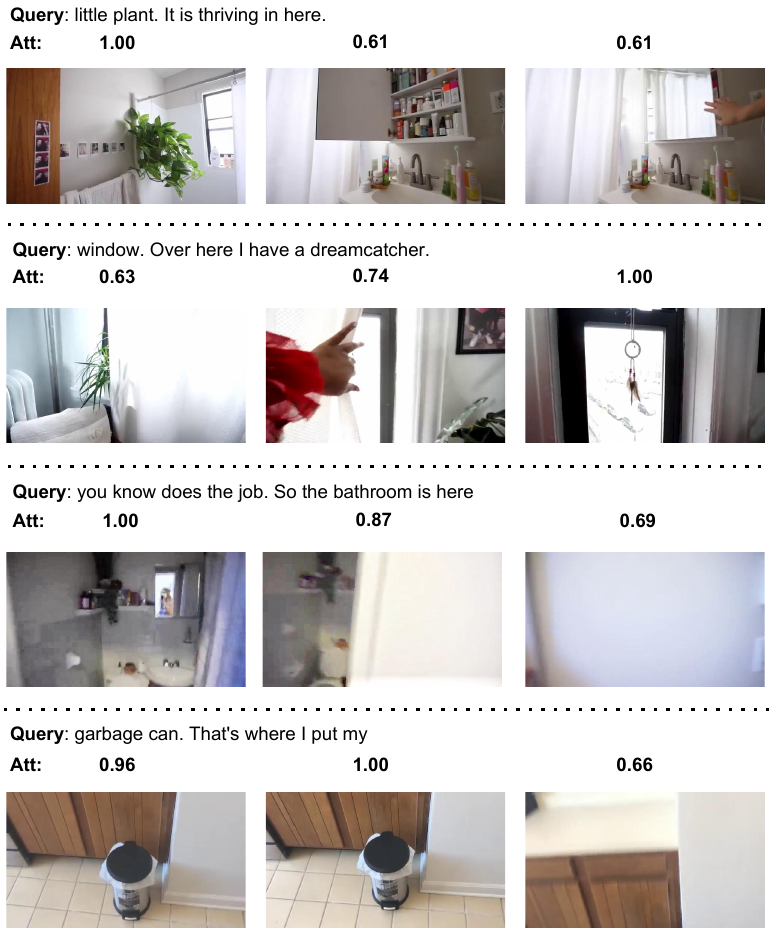}
\caption{ { Examples of the sentence-frame attention.} 
Each row contains a sentence-video pair. In each example, the query sentence is the subtitle corresponding to the video clip. For better viewing, only three representative frames are displayed for each video clip.
The attention score, denoting how much the frame contributes to the final video feature, is shown above each frame.
Note that the absolute value is related to the number of frames in the video.
For better comparison, the attention scores are normalized such that the largest score is 1.}
\label{fig:result_supp}
\end{figure*}

In Figure~\ref{fig:result_supp}, we show some more examples of how the input sentence attends to different frames in the video clip based on the relevance between the two modalities. 
The attention mechanism performs quite well in general. For instance, the first frame on the first row in Figure~\ref{fig:result_supp} scores the highest as it contains the plant mentioned in the query sentence, despite that the 3 frames in the video have a very similar background.

\begin{figure*}[t!]
\centering
\includegraphics[width=0.65\linewidth]{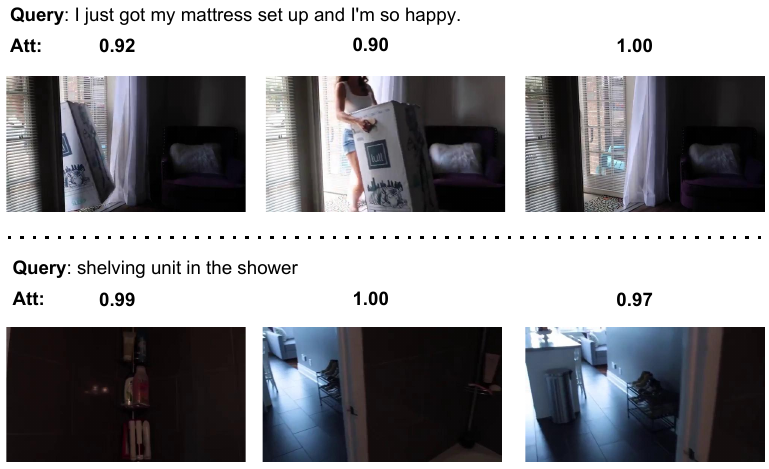}
\caption{ { Failure examples of the sentence-frame attention.} 
For better comparison, the attention scores are normalized such that the largest score is 1 in each example.}
\label{fig:result_fail}
\end{figure*}

There are also some rare cases where the attention mechanism fails.  Figure~\ref{fig:result_fail} shows two failure examples. 
In the first example, the network fails to attend to the mattress held by the woman. However, it might because that the mattress is folded and looks very different to what it usually does, \ie laid on the ground. That, therefore, could make it difficult for the network to recognize the mattress.
In the second case, the network fails to capture the shelving unit in the first frame. We think that it is due to the darkness of the image, and even human find it difficult to recognize.

\bibliographystyle{plain}
\bibliography{mybib}

\end{document}